\pdfoutput=1

\documentclass[11pt]{article}
\usepackage{float}

\usepackage[preprint]{acl}
\usepackage{listings}
\usepackage{adjustbox}
\usepackage{times}
\usepackage{xcolor}
\definecolor{SkyBlue}{rgb}{0.53, 0.81, 0.92}
\usepackage{multicol}

\usepackage{latexsym}
\usepackage{listings}
\usepackage{minted}  
\usepackage[utf8]{inputenc}
\usepackage{enumitem}
\usepackage{subcaption}
\usepackage{amsmath}
\usepackage{bm}
\usepackage{svg}
\usepackage{colortbl}        
\usepackage{booktabs}        
\usepackage{tabularx}
\usepackage{booktabs}
\usepackage{multirow}
\usepackage[T1]{fontenc}

\usepackage[utf8]{inputenc}

\usepackage{microtype}

\usepackage{inconsolata}

\usepackage{graphicx}

\hypersetup{colorlinks}

\usepackage[capitalize]{cleveref}
\crefname{section}{Sec.}{Secs.}
\Crefname{section}{Section}{Sections}
\Crefname{table}{Table}{Tables}
\crefname{table}{Tab.}{Tabs.}
\crefname{equation}{Eq.}{Eqs.}
\crefname{lstlisting}{Listing}{Listings}
\title{Context is Enough: Empirical Validation of \textit{Sequentiality} on Essays}

\author{Amal Sunny \\
  IIIT-Hyderabad \\
  \texttt{amal.sunny@research.iiit.ac.in} \\\And
  Advay Gupta \\
  IIIT-Hyderabad \\
  \texttt{advay.gupta@research.iiit.ac.in} \\\AND
  Vishnu Sreekumar \\
  IIIT-Hyderabad \\
  \texttt{vishnu.sreekumar@iiit.ac.in} }

\begin{document}
\maketitle
\begin{abstract}
Recent work has proposed using Large Language Models (LLMs) to quantify narrative flow through a measure called sequentiality, which combines topic and contextual terms. A recent critique argued that the original results were confounded by how topics were selected for the topic-based component, and noted that the metric had not been validated against ground-truth measures of flow. That work proposed using only the contextual term as a more conceptually valid and interpretable alternative. In this paper, we empirically validate that proposal. Using two essay datasets with human-annotated trait scores, ASAP++ and ELLIPSE, we show that the contextual version of sequentiality aligns more closely with human assessments of discourse-level traits such as Organization and Cohesion. While zero-shot prompted LLMs predict trait scores more accurately than the contextual measure alone, the contextual measure adds more predictive value than both the topic-only and original sequentiality formulations when combined with standard linguistic features. Notably, this combination also outperforms the zero-shot LLM predictions, highlighting the value of explicitly modeling sentence-to-sentence flow. Our findings support the use of context-based sequentiality as a validated, interpretable, and complementary feature for automated essay scoring and related NLP tasks.
\end{abstract}

\section{Introduction}
Large Language Models (LLMs) and measures derived from them have been used to investigate human behavior and cognition \cite{demszky2023using, mihalcea2024developments}. However, despite their growing use in cognitive science and education research, LLM-derived measures require careful validation to avoid misinterpretation and bias. In the context of narrative understanding, \citet{sap2022narrative} introduced the concept of \textit{sequentiality}, a measure combining a topic-based term and a contextual term to quantify narrative flow. A recent critique \cite{sunny_gupta_chandak_sreekumar_2025} showed that the topic term introduced a confound: topics used in its computation were drawn only from autobiographical narratives, biasing comparisons across story types. When topics were sampled uniformly across story types, the reported differences diminished substantially. That work proposed a simplified, contextual version of the metric and called for empirical validation using data with human annotations. In this paper, we test and validate both components of the sequentiality formula using two essay datasets, ASAP++ \cite{mathias-bhattacharyya-2018-asap} and ELLIPSE \cite{crossley2023english}, that include trait-level human scores for features aligned with narrative structure, such as Organization and Cohesion. Our results confirm that the contextual term captures narrative flow as previously hypothesized and that prior conclusions based on the full metric were likely affected by methodological artifacts.

To examine the validity of the contextual metric, \citet{sunny_gupta_chandak_sreekumar_2025} conducted a preliminary test using LLM-generated paragraphs with intentionally high and low narrative flow. The original sequentiality score failed to differentiate between them, while the contextual version captured the difference. These findings suggest that narrative flow is better described by how each sentence builds on its preceding context rather than by its topical similarity to a fixed set of story prompts.

To evaluate this hypothesis more rigorously, we use the ASAP++ and ELLIPSE datasets, which contain essays written in response to diverse prompts and annotated by human raters along traits such as Organization and Cohesion. These traits assess logical progression, sentence-level connectivity, and thematic development, which are elements that are conceptually linked to narrative flow. Both datasets have been widely used, with a few prior works focusing specifically on these trait-level scores \cite{doi-etal-2024-automated, ding-etal-2024-argumentation}. However, no previous study has evaluated LLM-based sequentiality measures against these human-annotated traits.

In addition to sequentiality, we consider linguistic features commonly used in automated essay scoring, such as word/sentence count and number of lemmas \cite{hou2025improve}. These features are known to correlate with general language abilities and content knowledge and have been effective in previous essay scoring models. However, they are only indirectly related to traits like Organization and Coherence, which depend on sentence-to-sentence flow. We evaluate whether sequentiality-based measures provide added predictive value beyond these linguistic baselines. Specifically, we compare the performance boost offered by the original sequentiality metric, its contextual variant, and its topic-only variant when combined with these features. This analysis addresses whether sequentiality captures distinct, conceptually relevant information about discourse flow that is not explained by surface-level linguistic features.

Our contributions in this study are threefold: 1) We validate the claim made by \citet{sunny_gupta_chandak_sreekumar_2025} that the formulation of \citeposs{sap2022narrative} sequentiality is flawed, and show that using only the contextual term achieves the best fit with human-annotated trait scores; 2) We assess the performance of the corrected sequentiality measure against direct zero-shot prompting of an LLM for flow-related trait scores, and show that while the metric captures aspects of flow, it is not optimized for the target trait in the way the prompted LLM response is; 3) We demonstrate that the inclusion of the contextual term provides a significant performance boost over established linguistic features, and does so more effectively than both the topic-only term and the original sequentiality formulation, suggesting that explicit modeling of sentence-to-sentence flow using only the contextual term improves predictive accuracy in automated essay scoring (AES) applications.

In the remainder of the paper, we describe the datasets and annotation schemes and formalize the sequentiality metrics. We then evaluate the original and reduced formulations against human ratings, compare them to established linguistic features, and assess whether the contextual term improves trait prediction in AES settings.

\section{Methods}

\subsection{Datasets}
We conduct experiments on two datasets, ELLIPSE and ASAP++. The ELLIPSE Corpus \cite{crossley2023english} consists of 6,483 essays across 44 unique prompts. Each essay has been annotated on 6 analytical traits, of which we consider only ``Cohesion'' due to it being the closest trait to narrative flow. 
ASAP++ \cite{mathias-bhattacharyya-2018-asap} is another popular dataset with trait-level scores across six distinct prompts.  For our analyses, we focus on Prompts 1 and 2 (see \cref{sec:additional_dataset_details} for details on prompts and traits), comprising 1,783 and 1,800 scored essays, respectively. We exclude the remaining prompts because they require examinees to incorporate external source materials, introducing direct in‐text references that could confound our sequentiality metrics, and the topic part of the prompt becomes comparable in length to the essay for these prompts. We consider the trait ``Organization'' in this dataset as the closest match for narrative flow. ELLIPSE is a more challenging dataset for our analyses because the linguistic features used were tailor-made for ASAP++ \cite{hou2025improve}. Furthermore, ELLIPSE is the bigger dataset with a greater number of prompts compared to ASAP++. Additional details and scoring rubrics are provided in \cref{sec:additional_dataset_details}
 
\subsection{Sequentiality}
\citet{sap2022narrative} define sequentiality as the difference in negative log-likelihood (NLL) between two language model predictions: one conditioned on the topic alone ($\mathrm{NLL}_{\mathcal{T}}$), and one conditioned on both the topic and preceding context ($\mathrm{NLL}_{\mathcal{C}}$). For a sentence $s_i$, sequentiality is computed as

\begin{equation}
\begin{aligned}
\Delta\ell(s_i) 
&= -\frac{1}{|s_i|} \Big[
    \underbrace{\log p_{LM}(s_i \mid T)}_{\text{\scriptsize topic-driven}} 
    \;\\
&\quad\quad\quad -\; 
    \underbrace{\log p_{LM}(s_i \mid T, s_{0:i-1})}_{\text{\scriptsize contextual}}
\Big],
\end{aligned}
\label{eq:sequentiality}
\end{equation}

where $|s_i|$ is the number of tokens in the sentence, \(T\) is the topic, and \(s_{0:i-1}\) are the preceding sentences.  We define the sentence-level negative log-likelihood (NLL) under a language model (LM) as
\begin{equation} 
\mathrm{NLL}(s|C) = -\frac{1}{|s|}\sum_{t=1}^{|s|}\log p_{\mathrm{LM}}(w_t|C,w_{0:t-1}), 
\end{equation}

where \(|s|\) is the length of \(s\), \(w_t\) is its \(t\)-th token, and  \(C\) is contextual conditioning (i.e., topic \(T\) and preceding sentences). The sequentiality of a paragraph is the mean NLL across all sentences in that paragraph.  Higher sequentiality values indicate that sentences are more predictable given the evolving context and fixed topic, whereas lower values signal greater divergence from the expectations set by the preceding text.

To isolate the influence of topic versus local context, we consider 
\(\mathrm{NLL}_{\mathcal{T}}\) and \(\mathrm{NLL}_{\mathcal{C}}\) separately to predict trait scores. We use the same 3 language models used in \citet{sunny_gupta_chandak_sreekumar_2025} to calculate these NLLs and sequentiality: LLaMa-3.1-8b-Instruct-AWQ (Llama; \citet{grattafiori2024llama}), Falcon3-10b-Instruct-AWQ (Falcon; \citet{Falcon3}) and Qwen-2.5-7b-Instruct-AWQ (Qwen; \citet{yang2024qwen2}). Please see \cref{sec:model_implementation} for details.

\subsection{Validation Methodology}

\subsubsection{Ordinal regression and model selection}
\label{sec:ordinal_reg_model}

The sequentiality measure, a composite metric introduced by \citet{sap2022narrative}, was proposed without empirical validation. To evaluate both its overall formulation and individual components, we compare them against trait-level human-annotated essay scores. Since the essay scores are ordinal and the sequentiality scores are continuous, we apply ordinal regression models where each component of the sequentiality term serves as a predictor and the human scores serve as the outcome variable.

We examine four models: a contextual model ($\mathrm{NLL}_{\mathcal{C}}$), a topic-based model ( $\mathrm{NLL}_{\mathcal{T}}$), a combined model ($\mathrm{NLL}_{\mathcal{C}}$ and $\mathrm{NLL}_{\mathcal{T}}$ as separate predictors), and a sequentiality model (the composite metric in \cref{eq:sequentiality}). Model fit was evaluated using the Akaike Information Criterion (AIC), with lower values indicating better performance. 
\subsubsection{Zero-shot prompted LLM}
We select the best-fitting measure based on AIC and assess its ability to predict essay trait scores compared to directly prompting an LLM. Prompts were prepared based on the rubric given to the human annotators for each dataset (see \cref{sec:prompt_scoring} for the prompts). We prompted the best model from \citet{sunny_gupta_chandak_sreekumar_2025}, Llama, to generate trait scores. 

We evaluate the results via a five-fold cross-validation on both datasets and report the mean Quadratic Weighted Kappa (QWK), in line with prior works \cite{taghipour-ng-2016-neural, ke-automated-survay-2019}.

\subsubsection{Incorporating linguistic features} 
Even if the contextual sequentiality term captures sentence-to-sentence flow, traits like cohesion and organization may not fully be captured by it. Further dimensions of the text may be required to fully explain these, for which we seek to incorporate some simple linguistic features with the goal of assessing the added benefit of sequentiality over and beyond these linguistic features. Based on \citet{hou2025improve}, we incorporate the 10 most impactful features in essay grading (see \cref{sec:ling_features} for a detailed list). We compare the trait prediction performance of regression models with just the linguistic features and those with variants of sequentiality (described in \cref{sec:ordinal_reg_model}) added to the linguistic features.

\section{Results}

\subsection{Comparing contextual and topic-driven terms in sequentiality}

\begin{table}[]
\centering
\resizebox{\columnwidth}{!}{%
\begin{tabular}{ccccc}
\toprule
\multirow{2}{*}{Model}  & \multirow{2}{*}{Features} &  \multicolumn{1}{c}{ASAP\_P1}       & \multicolumn{1}{c}{ASAP\_P2}       & \multicolumn{1}{c}{ELLIPSE} \\ \cline{3-5} 
                        &                                 & \multicolumn{2}{c}{Organization}                                       & Cohesion                     \\ \hline
\multirow{4}{*}{Llama}  & Seq                             & \multicolumn{1}{l}{4833}          & \multicolumn{1}{l}{5452}        & 21871                  \\  
                        & Topic                           & \multicolumn{1}{l}{4823}          & \multicolumn{1}{l}{5376}          & 21477                        \\  
                        & Context                         & \multicolumn{1}{l}{4674}          & \multicolumn{1}{l}{\textbf{5199}} & \textbf{20790}               \\  
                        & Both                            & \multicolumn{1}{l}{\textbf{4616}} & \multicolumn{1}{l}{\textbf{5179}} & \textbf{20790}               \\ \hline
\multirow{4}{*}{Qwen}   & Seq                             & \multicolumn{1}{l}{4839}          & \multicolumn{1}{l}{5454}        & 21870                  \\  
                        & Topic                           & \multicolumn{1}{l}{4821}          & \multicolumn{1}{l}{5348}          & 21430                        \\  
                        & Context                         & \multicolumn{1}{l}{4655}          & \multicolumn{1}{l}{\textbf{5151}} & \textbf{20690}               \\  
                        & Both                            & \multicolumn{1}{l}{\textbf{4593}} & \multicolumn{1}{l}{\textbf{5138}} & \textbf{20689}               \\ \hline
\multirow{4}{*}{Falcon} & Seq                             & \multicolumn{1}{l}{4820}          & \multicolumn{1}{l}{5450}      & 21874                    \\  
                        & Topic                           & \multicolumn{1}{l}{4828}          & \multicolumn{1}{l}{5361}          & 21337                        \\  
                        & Context                         & \multicolumn{1}{l}{4650}          & \multicolumn{1}{l}{\textbf{5131}} & \textbf{20674}               \\  
                        & Both                            & \multicolumn{1}{l}{\textbf{4560}} & \multicolumn{1}{l}{\textbf{5099}} & \textbf{20676}               \\ \hline
\end{tabular}%
}
\caption{Model fit (AIC) of sequentiality variants for essay traits in ASAP++ and ELLIPSE. Lower is better. Bold indicates the lowest AIC (and second-lowest if within 50 units).}
\label{tab:results_terms_compare}
\vspace{-2mm}
\end{table}

\cref{tab:results_terms_compare} summarizes model fit (AIC) for sequentiality components against human-rated flow-related scores (see \cref{sec:additional_results_model_fit} for additional traits). Across both datasets, $\mathrm{NLL}_{\mathcal{C}}$ consistently outperforms other terms, indicating better fit than either $\mathrm{NLL}_{\mathcal{T}}$ or full sequentiality. On ASAP, the combined model slightly outperforms $\mathrm{NLL}_{\mathcal{C}}$, but coefficient analysis reveals large differences from the original formulation: for ASAP, $(w_\mathcal{T} = -0.53, w_\mathcal{C} = 1.55)$; for ELLIPSE, $(w_\mathcal{T} = -0.004, w_\mathcal{C} = 1.23)$. These diverge from the original weights $(w_\mathcal{T} = -1.0, w_\mathcal{C} = 1.0)$ and support prior claims \cite{sunny_gupta_chandak_sreekumar_2025} that the context term captures flow and should drive any claimed differences in narrative flow (unlike in \citet{sap2022narrative}, where effects were driven by the topic term). Notably, on the larger ELLIPSE dataset, $\mathrm{NLL}_{\mathcal{T}}$ adds almost no value (AIC unchanged; $w_{\mathcal{T}} = -0.004$). Similar patterns hold across LLM backbones.

\subsection{Contextual term vs. zero-shot LLM}

\begin{figure}[t]
  \includegraphics[width=\columnwidth]{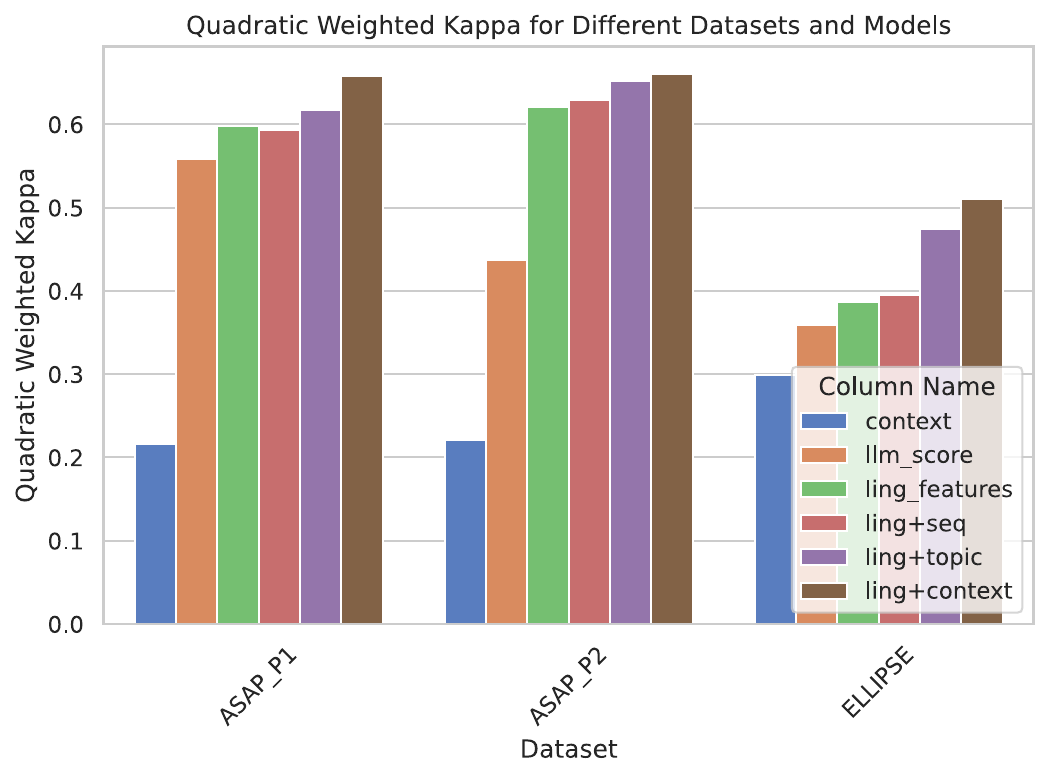}
  \caption{Comparison of QWK scores across approaches/features for each dataset’s flow-relevant trait (Organization for ASAP prompts 1 and 2, Cohesion for ELLIPSE), using Llama.}
  \label{fig:comparison_llmscore}
  \vspace{-2mm}
\end{figure}

Having demonstrated the limitations of the original sequentiality formulation, we now evaluate how well the contextual term ($\mathrm{NLL}_{\mathcal{C}}$) performs relative to a zero-shot prompted LLM. To do this, we prompt the LLM directly with the essay and scoring rubric to predict trait scores. This approach is conceptually distinct from sequentiality, as the LLM is explicitly asked to optimize for the human-defined scoring criteria. On the ASAP++ dataset, the LLM achieves a QWK score of 0.49, while on ELLIPSE it scores 0.36. These results are broadly consistent with prior work \cite{hou2025improve}, where prompted LLMs were shown to perform competitively on essay scoring tasks.
\looseness=-1
The contextual sequentiality measure, while conceptually aligned with discourse flow, performs worse than the prompted LLM (llm\_score in \cref{fig:comparison_llmscore}). This suggests that, as a standalone metric, $\mathrm{NLL}_{\mathcal{C}}$ may only partially capture the scoring rubric, potentially aligning with subcomponents like logical progression but not others such as completeness or relevance. Nevertheless, as we demonstrate next, this does not imply that sequentiality lacks utility as a feature in essay scoring.

\subsection{Additive effect of context and linguistic features}
To evaluate whether $\mathrm{NLL}_{\mathcal{C}}$ adds predictive value beyond existing baselines, we incorporate it into a standard feature-based AES model. As shown in prior work \cite{hou2025improve}, simple linguistic features such as word count and lemma diversity are effective predictors of essay quality. We adopt a similar set of features (detailed in \cref{sec:ling_features}) as our baseline model. Notably, these features alone (ling\_features in \cref{fig:comparison_llmscore}) already outperform the zero-shot LLM on both datasets, though the performance gap narrows on the more challenging ELLIPSE dataset.

We then augment this baseline with the different sequentiality components: the topic-only term ($\mathrm{NLL}_{\mathcal{T}}$), the contextual term ($\mathrm{NLL}_{\mathcal{C}}$), and the original sequentiality (\cref{eq:sequentiality}). As shown in \cref{fig:comparison_llmscore}, all three offer improvements over the linguistic baseline, but the gains from $\mathrm{NLL}_{\mathcal{C}}$ are consistently the largest. This further supports our claims and those of \citet{sunny_gupta_chandak_sreekumar_2025}, confirming that the contextual term is the most effective component of the sequentiality formulation for predicting human-annotated essay trait scores that are conceptually related to narrative flow. Additional results across other traits are provided in \cref{sec:additional_results_traits}, showing that the contextual term boosts performance on all traits and the overall score, consistent with prior suggestions that coherence measures are the most predictive of overall essay quality \cite{Crossley2010-CROCCA-3, crossley2011textCoherence}. $\mathrm{NLL}_{\mathcal{C}}$, therefore, is a quantitative measure that conceptually aligns more directly with human estimations of organization, coherence, and essay quality than surface-level linguistic features that are only indirectly (but strongly) indicative of essay quality.

\section{Conclusion}
We empirically validated the critique that the original sequentiality formulation is flawed due to its topic component \cite{sunny_gupta_chandak_sreekumar_2025}. Using trait-annotated essay datasets, we confirm that the context-only variant aligns better with human judgments of narrative flow. When used as a feature in automated essay scoring, this measure significantly improves predictive performance, highlighting the value of modeling sentence-to-sentence flow. These findings reinforce prior work \cite{Crossley2010-CROCCA-3, crossley2011textCoherence} showing that local coherence is central to perceived essay quality.

\section*{Limitations}

First, we rely on open-source mid-sized LLMs (LLaMa-3.1-7B, Falcon3-10B, and Qwen-2.5-7B) due to computational constraints. These models outperform GPT-3—the largest model evaluated in \citet{sap2022narrative} and represent a substantial improvement in reasoning and fluency. However, they may still lag behind the most capable proprietary models (e.g., GPT-4 or Claude 3), which could potentially offer even stronger contextual modeling. As such, our results may underestimate the full potential of sequentiality-based measures when used with the strongest available models.

Second, although we selected traits that are conceptually aligned with narrative flow (e.g., Organization, Cohesion), we observed performance gains across all traits and overall essay scores when including the context-based sequentiality term. While this is consistent with known correlations between trait scores that are often rated by the same human annotators within each dataset, it limits the specificity of our claims. That is, we cannot definitively conclude that context-sequentiality captures only sentence-to-sentence flow rather than broader indicators of essay quality. Future work should isolate this effect more clearly, for instance by comparing the impact of context-sequentiality against other discourse features or using targeted human judgments of local coherence and flow.

\section*{Ethics Statement}
The datasets used in this study are publicly available and widely used for academic research in automated essay scoring. Both datasets are anonymized, with ASAP++ explicitly removing identifiable information. All model inference was performed using open-source LLMs run locally, eliminating the risk of data leakage and minimizing concerns around privacy. 

An additional ethical consideration is that features derived from LLMs may encode biases present in their pretraining data. As a result, these features could favor certain writing styles or discourse patterns, potentially reinforcing normative preferences that disadvantage other valid forms of expression. Awareness of such biases is important when integrating LLM-derived measures into AES systems.


\bibliography{custom}

\appendix
\section{Dataset details}
\label{sec:additional_dataset_details}
\subsection{ASAP++ dataset}
The dataset has 6 prompts; however, prompts 3-6 use texts from an external source and have in-text references which may potentially confound NLL calculations, thus we only use prompts 1 and 2 which can be found in \cref{lst:prompts_asap}.

\begin{lstlisting}[caption={Prompts 1 and 2 from the ASAP++ dataset}, label={lst:prompts_asap}]
Prompt 1: 
More and more people use computers, but not everyone agrees that this benefits society. Those who support advances in technology believe that computers have a positive effect on people. They teach hand-eye coordination, give people the ability to learn about faraway places and people, and even allow people to talk online with other people. Others have different ideas. Some experts are concerned that people are spending too much time on their computers and less time exercising, enjoying nature, and interacting with family and friends.
Write a letter to your local newspaper in which you state your opinion on the effects computers have on people. Persuade the readers to agree with you.

Prompt 2:
"All of us can think of a book that we hope none of our children or any other children have taken off the shelf. But if I have the right to remove that book from the shelf -- that work I abhor -- then you also have exactly the same right and so does everyone else. And then we have no books left on the shelf for any of us." --Katherine Paterson, Author
Write a persuasive essay to a newspaper reflecting your views on censorship in libraries. Do you believe that certain materials, such as books, music, movies, magazines, etc., should be removed from the shelves if they are found offensive? Support your position with convincing arguments from your own experience, observations, and/or reading.
\end{lstlisting}

For prompts 1 and 2, each essay is rated on the following five attributes:

\begin{enumerate}
\item \textbf{Ideas \& Content}: The amount of content and ideas present in the essay.
\item \textbf{Organization}: How well structured the essay is.
\item \textbf{Word Choice}: Precision, variety, and impact of vocabulary.
\item \textbf{Sentence Fluency}: Rhythm, variation, and smoothness of sentences.
\item \textbf{Conventions}: Correctness of punctuation, spelling, grammar, and usage.
\end{enumerate}

\subsection{ELLIPSE dataset}
Each essay receives a holistic score reflecting overall proficiency and communicative effectiveness. It is also scored across the following analytic dimensions:

\begin{enumerate}
\item \textbf{Cohesion}: Text organization and use of cohesive devices.
\item \textbf{Syntax}: Range and control of sentence structures.
\item \textbf{Vocabulary}: Precision, range, and appropriateness of word choice.
\item \textbf{Phraseology}: Use of idioms, collocations, and lexical bundles.
\item \textbf{Grammar}: Control of morphology and grammatical usage.
\item \textbf{Conventions}: Spelling, punctuation, and capitalization.
\end{enumerate}
\subsection{Scoring rubric}
\label{sec:scoring_rubric}

The rubric for the ``Cohesion'' trait in the ELLIPSE dataset is given in \cref{lst:cohesion_rub} and  for the ``Organization'' trait in the ASAP++ dataset can be found in \cref{lst:organization_rub}.
\begin{lstlisting}[caption={Cohesion rubric},label={lst:cohesion_rub}]
Cohesion
This property checks how well structured the essay is. 

Score 5: Text organization consistently well controlled using a variety of effective  linguistic features such  as reference and transitional words and phrases to connect ideas across sentences and paragraphs; appropriate  overlap of ideas.

Score 4: Organization generally well controlled; a range of cohesive devices used appropriately such as reference and transitional words and phrases to connect ideas; generally appropriate overlap of ideas.

Score 3: Organization generally controlled; cohesive devices used but limited in type; Some repetitive, mechanical, or faulty use of cohesion use within and/or between sentences and paragraphs.

Score 2: Organization only partially developed with a lack of logical sequencing of ideas; some basic cohesive devices used but with inaccuracy or repetition.

Score 1: No clear control of organization; cohesive devices not present or unsuccessfully used; presentation of ideas unclear.
\end{lstlisting}

\begin{lstlisting}[caption={Organization rubric}, label={lst:organization_rub}]
Organization
This property checks how well structured the essay is. NOTE: Since the dataset has the essays compressed into one line, please bear in mind that the paragraph information is lost. Hence, give writers the benefit of the doubt here.

Score 6: The essay is well-organized. There is a clear flow of ideas with each idea self-contained (this is where we assume that each idea is contained in a paragraph). The essay has the appropriate form as a letter to the editor.

Score 5: The essay shows good organization. There is a flow of ideas. However, the ideas are mostly self-contained. The essay has the appropriate form as a letter to the editor. 

Score 4: The essay shows satisfactory organization. It contains a basic introduction, body and conclusion.

Score 3: The essay shows some organization. Its form may not be that of a letter to the editor. Its ideas are not necessarily self-contained.

Score 2: Shows little or no evidence of organization.

Score 1: The essay is awkward and fragmented. Ideas are not self-contained.
\end{lstlisting}

\section{LLM implementation }
\label{sec:model_implementation}
All models were implemented using \texttt{transformers} v4.46.3 \cite{wolf-etal-2020-transformers}. Sequentiality was computed using default parameters following \citet{sunny_gupta_chandak_sreekumar_2025}. For LLM scoring, we set the temperature to 0.0001 to ensure reproducibility, leaving other parameters at their defaults. Computation times for LLM scoring and sequentiality were approximately 24 and 40 hours, respectively, on a single RTX 2080Ti across all three datasets.

\section{Linguistic features}
\label{sec:ling_features}
\cref{tab:ling-features-list} summarizes the set of linguistic features used in this study. We follow \citet{hou2025improve}, who selected the top 10 non-prompt-specific features from state-of-the-art AES models \cite{Ridley_He_Dai_Huang_Chen_2021, li-ng-2024-conundrums}. Feature extraction follows the same pipeline: total word and sentence counts are computed using \texttt{nltk} \cite{bird-loper-2004-nltk}; lemma, noun, and stopword counts are obtained using \texttt{spaCy}’s \texttt{en\_core\_web\_sm} model \cite{honnibal-spacy-2020}; and Dale–Chall difficult word count and long word count are extracted using the \texttt{readability} library\footnote{\url{https://pypi.org/project/readability/}} based on \citet{dale1948formula}. See \cref{tab:ling-features-list} for feature definitions.

\begin{table*}[h!]
  \centering
  \begin{tabular}{p{4cm} p{10cm}}
    \hline
    \textbf{Feature} & \textbf{Definition} \\
    \hline
    Total number of unique words & Count of word types that occur exactly once in the essay; a proxy for lexical diversity. \\
    Total number of words & Total token count of the essay, indicating overall length and opportunity for argument development. \\
    Total number of sentences & Number of sentence boundaries detected, used to gauge sentence complexity (e.g., average words per sentence). \\
    Long word count &  Count of long words implemeted in the Readabiliity package \\
    Character count (non-space, non-punctuation) & Number of letters and digits only, excluding spaces and punctuation; aligns with some readability metrics. \\
    Character count (all characters) & Total count of all characters including spaces and punctuation; reflects full essay length as stored. \\
    Total number of lemmas & Count of base‐form tokens after lemmatization, measuring conceptual variety beyond inflected forms. \\
    Total number of nouns & Count of tokens tagged as nouns (POS = NN, NNS, etc.), indicating topical density and concreteness. \\
    Total number of stopwords & Count of high‐frequency function words (e.g., “the”, “is”, “and”), reflecting balance between function and content words. \\
    Dale–Chall difficult word count & Number of words not in the Dale–Chall list of 3,000 common words (understood by 80\% of fifth graders), measuring vocabulary challenge. \\
    
    \hline
  \end{tabular}
  \caption{Descriptions of the linguistic features.}
  \label{tab:ling-features-list}
\end{table*}

\section{LLM prompts}
\label{sec:prompt_scoring}
\cref{lst:cohesion_prompt} and \cref{lst:organization_prompt} show the LLM prompt templates used to get ``Cohesion'' and ``Organization'' scores, respectively. Each of these has the \texttt{\{rubric\}}, \texttt{\{topic\}} and \texttt{\{essay\}} replaced with their corresponding rubrics (the same rubrics provided to the annotators in \cref{sec:scoring_rubric}), essay and topics from the essay being scored.

\begin{lstlisting}[caption={Prompt template for Cohesion annotation}, label={lst:cohesion_prompt}]
You are an annotator highly competent in grading English essays. Your task is to grade the following essay, given the topic the essay was written on and a rubric to grade the essay. 

Rubric: {rubric} 
Topic: {topic} 
Essay: {essay} 
\end{lstlisting}

\begin{lstlisting}[caption={Prompt template for Organization annotation},label={lst:organization_prompt}]
You are an annotator highly competent in grading English essays. Your task is to grade the following essay, given the prompt used to write the essay and a rubric to grade the essay. Additionally consider that all the essays are anonymized. This means that the named entities (people, places, dates, times, organizations, etc.) are replaced by placeholders (Eg. @NAME1, @LOCATION1, etc.). In addition to this, capitalized phrases are anonymized as @CAP1, @CAP2, etc. These anonymizations should not affect your scoring. You are free to replace the anonymizations with any placeholders.

Rubric: {rubric}
Prompt: {topic}
Essay: {essay}
\end{lstlisting}

\section{Additional results}
\label{sec:additional_results}

\subsection{Model fit comparisons}
\label{sec:additional_results_model_fit}
\cref{tab:results_terms_compare_full} reports model fit comparisons across additional traits that may reflect narrative flow. The results mirror those observed for the primary traits: the contextual term alone generally provides the best fit, while the combined contextual+topic term yields marginal improvements on ASAP++, but shows negligible gains on the larger ELLIPSE dataset.

\begin{table*}[]
\centering
\resizebox{\textwidth}{!}{%
\begin{tabular}{@{}llllllllll@{}}
\toprule
\multirow{2}{*}{Model} & \multirow{2}{*}{Features} & \multicolumn{3}{c}{ASAP P1} & \multicolumn{3}{c}{ASAP P2} & \multicolumn{2}{c}{ELLIPSE} \\ \cmidrule(l){3-10} 
                        &         & Organization  & Sentence fluency & Content       & Organization  & Sentence fluency & Content       & Cohesion       & Overall        \\ \cmidrule(r){1-10}
\multirow{4}{*}{Llama}                   & Seq     & 4833          & 4919             & 4988          & 5452          & 5259             & 5600          & 21871          & 21277          \\
                        & Topic   & 4823          & 4854             & 4981          & 5376          & 5166             & 5509          & 21477          & 20626          \\
                        & Context & 4674          & \textbf{4718}    & \textbf{4876} & \textbf{5199} & \textbf{4959}    & \textbf{5345} & \textbf{20790} & \textbf{19714} \\
                        & Both    & \textbf{4616} & \textbf{4698}    & \textbf{4838} & \textbf{5179} & \textbf{4960}    & \textbf{5334} & \textbf{20790} & \textbf{19714} \\ \cmidrule(l){1-10} 
\multirow{4}{*}{Qwen}   & Seq     & 4839          & 4920           & 4992          & 5454        & 5249             & 5599          & 21870          & 21276          \\
                        & Topic   & 4821          & 4849             & 4979          & 5348          & 5077             & 5479          & 21430          & 20554          \\
                        & Context & 4655          & \textbf{4693}    & \textbf{4862} & \textbf{5151} & \textbf{4907}    & \textbf{5297} & \textbf{20690} & \textbf{19563} \\
                        & Both    & \textbf{4593} & \textbf{4670}    & \textbf{4820} & \textbf{5138} & \textbf{4909}    & \textbf{5291} & \textbf{20689} & \textbf{19563} \\ \cmidrule(l){1-10} 
\multirow{4}{*}{Falcon} & Seq     & 4820          & 4908             & 4972          & 5450      & 5261           & 5600          & 21874          & 21278          \\
                        & Topic   & 4828          & 4865             & 4985          & 5361          & 5094             & 5493          & 21337          & 20448          \\
                        & Context & 4650          & 4677             & 4849          & \textbf{5131} & \textbf{4884}    & \textbf{5283} & \textbf{20674} & \textbf{19550} \\
                        & Both    & \textbf{4560} & \textbf{4622}    & \textbf{4776} & \textbf{5099} & \textbf{4882}    & \textbf{5264} & \textbf{20676} & \textbf{19549} \\ \cmidrule(l){1-10} 
\end{tabular}%
}
\caption{Model fit (AIC) comparison of sequentiality variants—$\mathrm{NLL}_{\mathcal{T}}$, $\mathrm{NLL}_{\mathcal{C}}$, and their combination—across an extended list of essay traits in ASAP++ and ELLIPSE. Each cell reports the AIC for models using the specified features. Lower values indicate better fit. Bold indicates the lowest AIC (and second-lowest if within 50 units).}
\label{tab:results_terms_compare_full}
\end{table*}

\subsection{Additional trait-level performance comparisons}
\label{sec:additional_results_traits}
\begin{figure*}[t]
  \includegraphics[width=\textwidth]{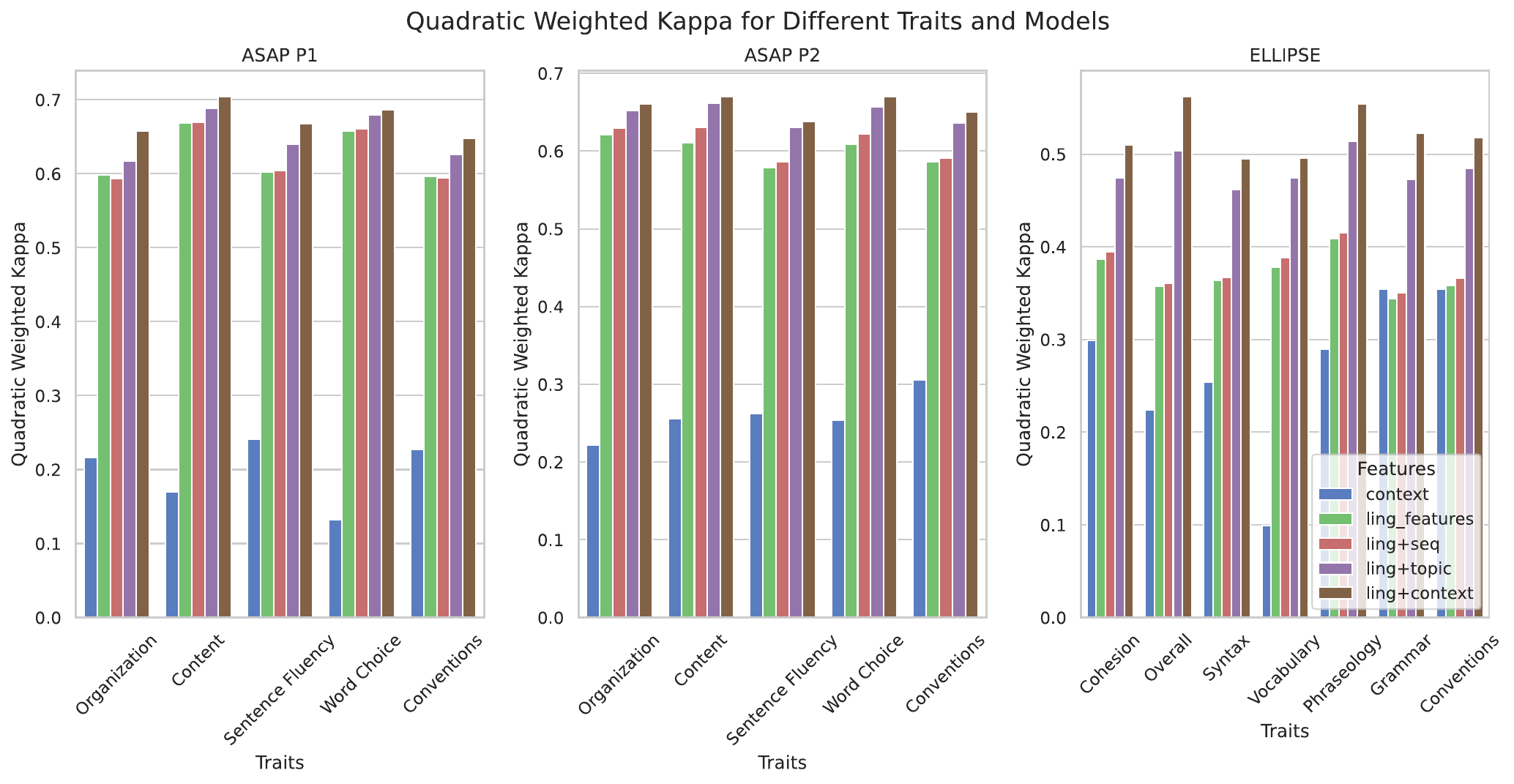}
  \caption{Comparison of QWK scores across approaches/features for all traits for both datasets, using LLaMa.}
  \label{fig:comparison_all_traits}
\end{figure*}

\cref{fig:comparison_all_traits} shows performance comparisons across different input features and trait scores. Overall, the trends align with those observed in the main analysis: the contextual term underperforms direct LLM scoring, linguistic features alone outperform the LLM, and the inclusion of the contextual term yields the largest performance gain when added to linguistic features. Notably, these improvements extend even to traits not directly related to narrative flow. One possible explanation is that the sequentiality measure captures broader aspects of essay quality, which are correlated with multiple trait scores, thereby improving predictions across a wider range of dimensions.

\end{document}